\documentclass[letterpaper, 10 pt, conference]{ieeeconf}  

\IEEEoverridecommandlockouts                              

\usepackage{cite}
\usepackage{amsmath,amssymb,amsfonts}
\usepackage{graphicx}
\usepackage{textcomp}
\usepackage{xcolor}
\usepackage{url}
\usepackage{tablefootnote}

\usepackage{algorithm}
\usepackage{algpseudocode}
\usepackage{subcaption}
\usepackage{multirow}
\usepackage{booktabs}
\usepackage{makecell}

\usepackage{booktabs}
\usepackage{hyperref}

\newcommand{\modelname}{\text{STR2}}
\newcommand{\modelnamespace}{\text{STR2} }

\title{\LARGE \bf
Generalizing Motion Planners with Mixture of Experts for Autonomous Driving
}

\author{Qiao Sun$^{1\hspace{1pt}*}$\thanks{\hspace{1pt}*Denotes equal contribution.}, 
Huimin Wang$^{4\hspace{1pt}*}$,
Jiahao Zhan$^{2}$, Fan Nie$^{3}$, 
Xin Wen$^{4}$, Leimeng Xu$^{4}$, \\
Kun Zhan$^{4}$, Peng Jia$^{4}$, Xianpeng Lang$^{4}$
Hang Zhao$^{1,5}$
\thanks{$^{1}$Shanghai Qi Zhi Institute $^{2}$Fudan University $^{3}$Stanford University $^{4}$LiAuto $^{5}$Tsinghua University}
\thanks{Corresponding at: {\texttt{hangzhao@mail.tsinghua.edu.cn}}}
}

\begin{document}

\maketitle
\thispagestyle{empty}
\pagestyle{empty}

\begin{abstract}

Large real-world driving datasets have sparked significant research into various aspects of data-driven motion planners for autonomous driving. These include data augmentation, model architecture, reward design, training strategies, and planner pipelines. 
These planners promise better generalizations on complicated and few-shot cases than previous methods.
However, experiment results show that many of these approaches produce limited generalization abilities in planning performance due to overly complex designs or training paradigms. 
In this paper, we review and benchmark previous methods focusing on generalizations. The experimental results indicate that as models are appropriately scaled, many design elements become redundant. We introduce StateTransformer-2 (\modelname), a scalable, decoder-only motion planner that uses a Vision Transformer (ViT) encoder and a mixture-of-experts (MoE) causal Transformer architecture. The MoE backbone addresses modality collapse and reward balancing by expert routing during training. Extensive experiments on the NuPlan dataset show that our method generalizes better than previous approaches across different test sets and closed-loop simulations. Furthermore, we assess its scalability on billions of real-world urban driving scenarios, demonstrating consistent accuracy improvements as both data and model size grow.


\end{abstract}

\section{INTRODUCTION}


Generalization bottlenecks the performance of autonomous driving motion planning for complex cases in the real world. These challenges arise from inconsistent objectives in complex environments. For example, human drivers might cross solid white lines while overtaking slow traffic ahead. Learning-based planners offer promising solutions by learning the complex mapping between observations and driving decisions from large datasets or simulations.
A generalizable policy needs to react to similar scenarios and balance different objectives when scenarios get complicated.
Considering the recent advances in the generalization of large language and vision models, scaling learning-based motion planners, including the training set and model sizes, could solve complicated, few-shot, and zero-shot driving problems. Additionally, MoE \cite{jiang2024mixtral} Transformers architectures better learn and balance complex preferences. 

\begin{figure}[t]
    \centering
    \includegraphics[width=0.48\textwidth]{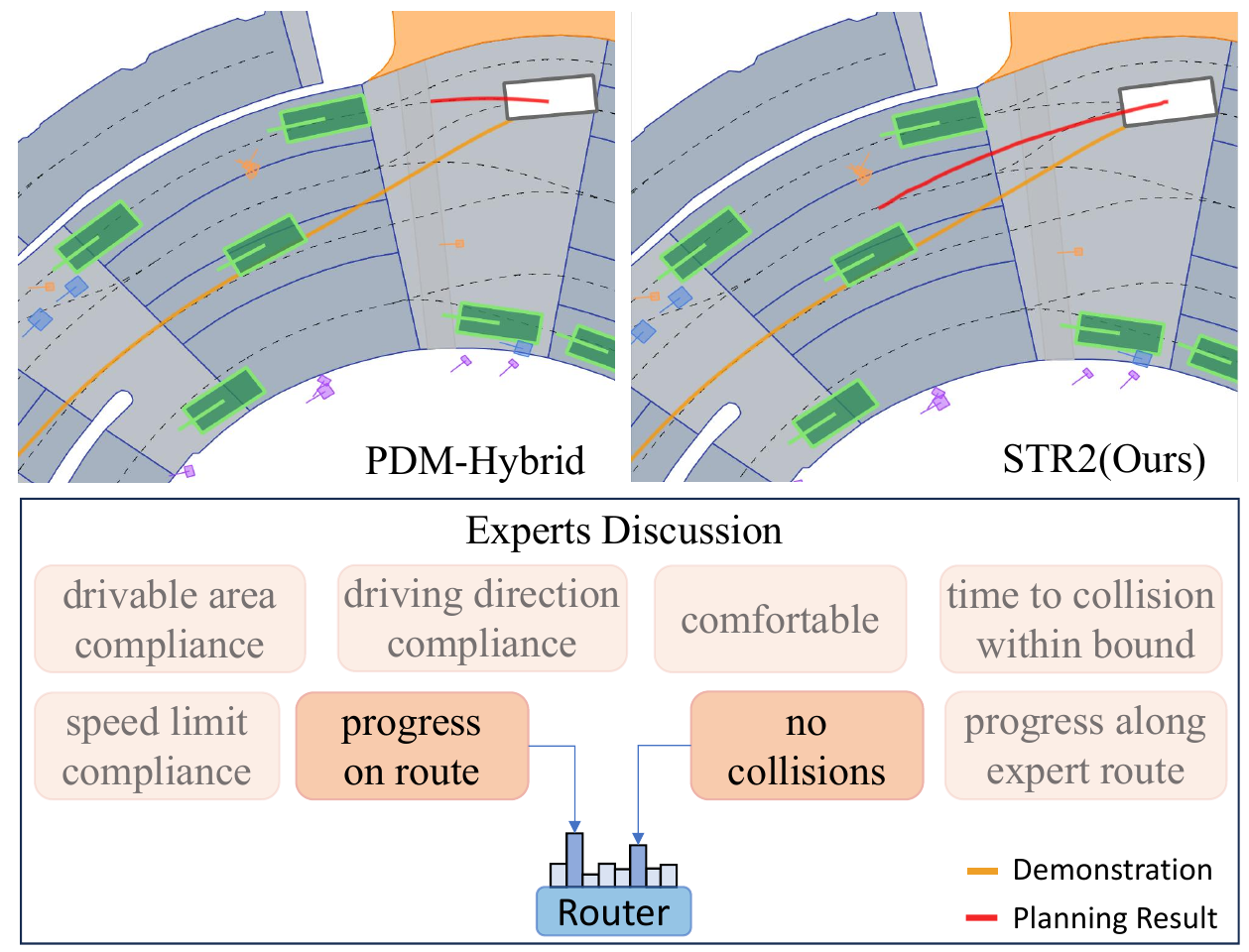}
    \caption{
    The planning results, in red, from PDM-Hybrid and \modelnamespace at the pickup area at the top, and an illustration of the MoE model learning and balancing different explicit rewards at the bottom.
    In this case, \modelnamespace produces a better-nudging trajectory by balancing two conflicting rewards, making progress, and avoiding collisions.
    }
    \label{fig:teaser}
\vspace{-6mm}
\end{figure}

Scaling for generalizations requires a large, diverse, and high-quality dataset for training and testing. 
We select the NuPlan \cite{caesar2021nuplan} dataset for training, which offers a much larger and more comprehensive collection of diverse urban driving scenarios compared to previous datasets~\cite{ettinger2021large,caesar2020nuscenes,Argoverse2}. 
For testing, ensuring a sufficiently large test set is critical for testing generalities, especially for evaluating complicated cases. 
However, testing motion planners using closed-loop simulations is resource-intensive and time-consuming due to slow CPU model inferences. As a result, no open test server is available, which has led previous methods \cite{dauner2023parting, cheng2024rethinking, huang2023dtpp} to propose their own preferred subset to report results, complicating direct comparison. 
To address this issue, we accelerate the simulation process by GPU-parallel model inferences in batches for better efficiency with larger test sets. 
We extract two larger 4k test sets from the validation and the test dataset and provide a distinctively more comprehensive performance benchmark by thoroughly testing previous methods with their official checkpoints against various subsets. 
On the other hand, testing generality with complicated and few-shot cases presents significant challenges due to the difficulty of simulating realistic reactions during closed-loop simulations.
We implement a wide spectrum of tests to benchmark the planning performance. 
First of all, closed-loop simulations challenge the models by drifting to different positions not visited in the training set. For example, models with low generalization abilities might not be able to correct themselves from the edge of the road because there are few or no samples at those dangerous positions in the training set for models to learn. 
Secondly, reactive simulations challenge the models with the environment agents' out-of-distribution reactions to potential conflicts.
Finally, unseen new synthetic scenarios from InterPlan \cite{hallgarten2024can}, like a crash site, challenge the zero-shot motion planning abilities of the models.

Our experiment results suggest that various previous methods greatly suffer from causality confusions, as mentioned in \cite{cheng2024rethinking}, producing over-smooth trajectories with modality collapse. 
Inspired by \cite{fedus2022switch}, we propose \modelnamespace to model different driving rewards by different experts via a router at each layer by an MoE backbone, as illustrated in Fig. \ref{fig:teaser}. 
On the other hand, inspired by multi-modality foundation models \cite{singh2022flava}, we formulate the motion planning task as a general sequence modeling task and scale our decoder-only MoE backbone \cite{jiang2024mixtral} up to 800 million parameters with a ViT \cite{dosovitskiy2020vit} encoder and a two-layer MLP decoder. 
Unlike some previous methods, we do not add additional training paradigms like reinforcement learning (RL) \cite{cheng2024rethinking}, inverse reinforcement learning (IRL) \cite{huang2023dtpp} or contrastive learning \cite{cheng2024pluto}. In contrast, we train \modelnamespace with a straightforward one-stage self-supervised learning without additional reward engineering for the testing metrics. We implement an autoregressive process for efficient training and flexible sampling. Inspired by \cite{zhao2021tnt, shi2022motion}, we involve an additional proposal classification before generating key points and trajectories. The proposal classification further avoids modality collapse on the trajectory curvatures in regression on the continuous space during learning. 
As a result, comprehensive experiment results indicate a better performance on all metrics against all testing datasets than previous methods with a more general but more challenging raster representation of the environment.
Additionally, our method produces much smaller performance drops when tested with reactive closed-loop simulations, as well as with unseen scenarios than other previous methods.

In summary, our contributions are:
\begin{itemize}
    \item We propose a scalable MoE-based autoregressive model to learn different explicit rewards for motion planning and our method outperforms previous state-of-the-art methods by scaling with self-supervisions.
    \item We comprehensively benchmark and analyze the generality of previous methods across multiple test sets with multiple Closed-Loop metrics by speeding up the model inference during simulations.
    \item We unprecedentedly present scaling experiments with up to billions of diverse real-world urban driving scenarios.
    \item We release our codes for training, testing, and simulations on the NuPlan dataset for easy reproduction at \href{https://github.com/Tsinghua-MARS-Lab/StateTransformer}{https://github.com/Tsinghua-MARS-Lab/StateTransformer}.
\end{itemize}

\section{Related Work}

The non-data-driven planning systems are widely used in academia and industry \cite{paden2016survey} for their reliability in producing optimal solutions. However, as the testing environment gets more complicated, engineered cost functions are getting more difficult to cover the human-like balancing of simple objectives like comfort, safety, and efficiency. On the other hand, reinforcement learning (RL) methods also guarantee an optimal solution with generalization abilities but suffer from significant sim-to-real gaps when deployed on large-scale AV fleets.

\subsection{Imitation Learning  based Planners}

Imitation Learning (IL) methods learn driving policies from experienced drivers. These self-supervised learning models are easy to scale without additional data labeling and promise better performance as long as the size of the training data keeps growing. 
The challenge is that previous IL methods suffer from poor generalization abilities on motion planning tasks, often represented as covariate shift problems like accumulative error, due to complex environments and rewards.
PDM-Hybrid\cite{dauner2023parting} utilizes imitation loss to refine trajectories based on rule-based center lines.
GameFormer\cite{huang2023gameformer} hierarchically models the relationships between scene elements by query-based cross-attention mechanisms.
PlanTF\cite{cheng2024rethinking} explores different augmentation and dropout strategies to mitigate compounding errors from a weak model.
DTPP\cite{huang2024dtpp} employs a tree-structured policy planner and proposes a differentiable joint training framework for both ego-conditioned prediction and a cost model, which is trained by an inverse reinforcement learning (IRL) paradigm.
PLUTO\cite{cheng2024pluto} provides additional hand-crafted reward functions, as heat maps which can be dated back to \cite{bansal2018chauffeurnet, zhou2021exploring}. Reward engineering is dangerous because it can be simply another way of overfitting the dataset. 
GUMP \cite{hu2024solving} uses Soft Actor-Critic (SAC) \cite{haarnoja2018soft} with reinforcement learning (RL).
Sampling and searching are also useful techniques widely used by IL planners \cite{dauner2023parting, huang2023dtpp, cui2021lookout}, as well as self-supervised generative models on other tasks, like language modeling. 

\subsection{Scaling Laws}

Previous studies \cite{scalingopenai, hoffmann2022trainingcomputeoptimallargelanguage} provide adequate empirical results indicating a log-log relationship between the accuracy and size of the training dataset, as well as the accuracy and the model parameters on language modeling tasks. Considering these Transformer architectures are also good at modeling time series trajectories, it is intuitive to scale with them for better generalization abilities on the autonomous driving motion planning problem. Some early studies \cite{sun2023large, hu2024solving} suggest similar scaling properties by training with models similar to the language models.

\begin{figure*}[t]
    \centering
    \includegraphics[width=0.99\textwidth]{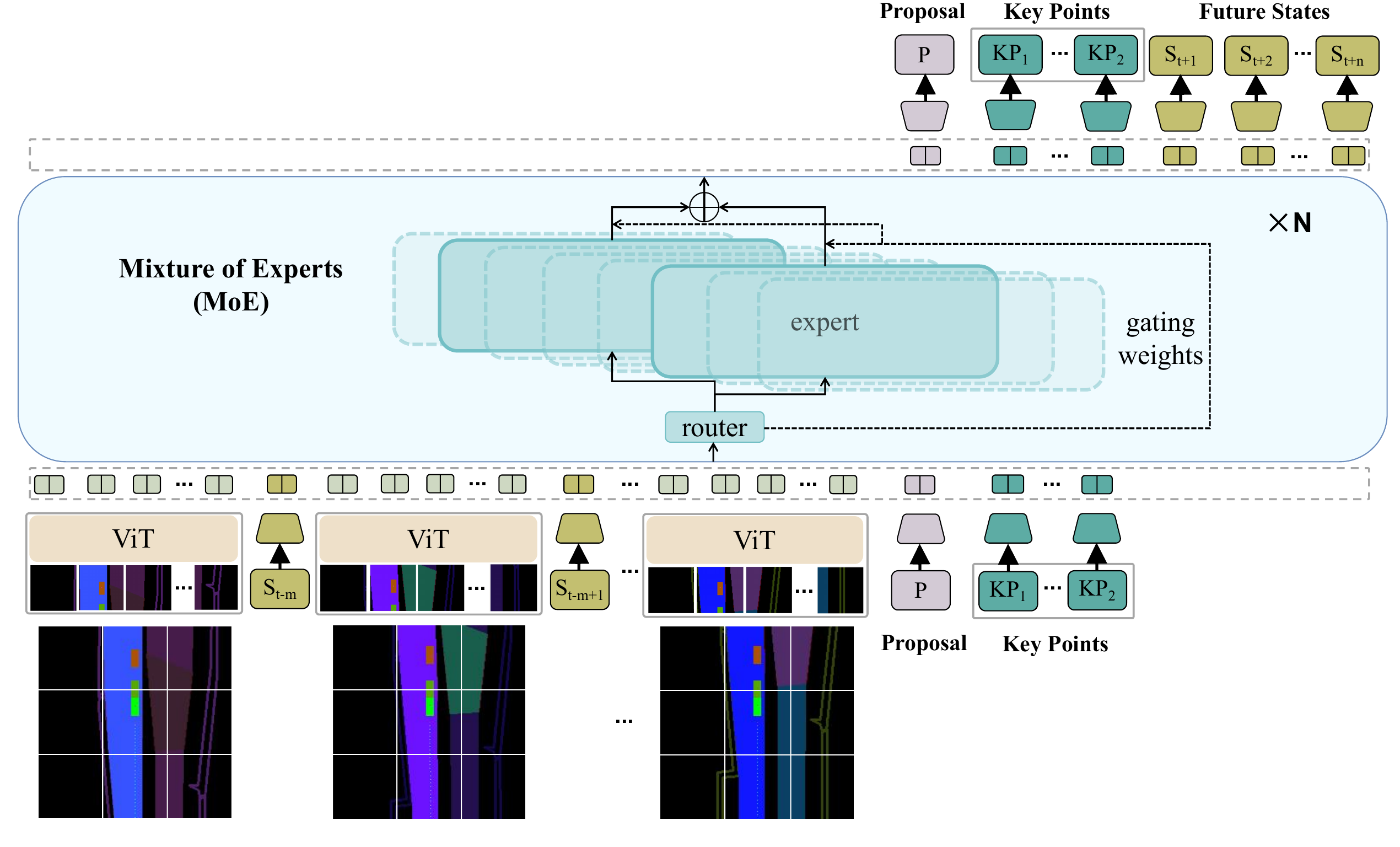}
    \caption{An overview of the \modelname-CPKS model which has a sequence of context, proposal, key points, and future states for the MoE backbone to model. For \modelname-CKS, proposals are removed in the sequence for better efficiency. The context part has rasterized environment information encoded by scalable ViT encoders and past ego states.}
    \label{fig:model}
\vspace{-4mm}
\end{figure*}

\subsection{Simulation and Testings}
NuPlan \cite{caesar2021nuplan} is a large-scale dataset aiming to fulfill the learning-based planners' needs of a huge training dataset. The dataset encompasses 1300 hours of recorded driving data collected from 4 urban centers, segmented into 75 scenario types using automated labeling tools. Although it is instinctive to benchmark different methods' performance by trajectory accuracy, which is the offset between model outputs and the ground truth trajectory, some \cite{li2024egostatusneedopenloop, dauner2023parting} argue a significant gap between the open-loop prediction performance and the effectiveness in the real world. However, it is challenging to test the performance within a simulation. Firstly, the controller might produce a tracking error \cite{cheng2024rethinking} as a systematic bias. Furthermore, it is hard \cite{hallgarten2024can} to simulate the reaction of other road users responding to the models' output. Without a perfect solution, we cover as many tests and metrics as possible to comprehensively benchmark each method. Finally, rule-based planners and reward engineering cost maps can easily overfit the test set by tuning against failed cases in the test set.

\section{Scaling for Generality with MoE}

Following previous learning-based motion planners, our model learns the mapping from observations of 2 past seconds to an optimal 8 seconds of future ego trajectory. To ensure scalability, we follow \cite{sun2023large} to formulate this task as a sequence modeling problem to learn with an autoregressive model. In this section, we will discuss details about data preprocessing, rasterizations, model designs, training details, the dataset for training and testing, the metrics we choose to evaluate, and the LiAuto dataset for scaling.

\begin{table}[h]
\caption{MoE Backbone Architecture}
\label{MoE_Backbone}
\begin{center}
\begin{tabular}{l c c c c}
\toprule
Parameter & MoE-100m & MoE-800m & MoE-1b\\
\midrule
dim  & 320 & 512 & 1024 \\
n\_layers & 16 & 32 & 16 \\
hidden\_dim & 1280 & 2048 & 4096 \\
n\_heads & 16 & 32 & 16 \\
n\_kv\_heads & 8 & 8 & 8 \\
num\_experts & 8 & 8 & 8 \\
top\_k\_experts & 2 & 2 & 2 \\
\bottomrule
\end{tabular}
\end{center}
\vspace{-6mm}
\end{table}

\subsection{Training and Testing Datasets}

\begin{table*}[t]
\caption{Performance comparison with various state-of-the-arts with NuPlan Closed-Loop Metrics. NR is the overall score of the closed-loop simulation with non-reactive agents. R is the overall score of the closed-loop simulation with reactive agents controlled by a rule-based planner.}
\vspace{-2mm}
\label{table:closedloop}
\begin{center}
\begin{tabular}{lcccccccccc}
\toprule
 & \multicolumn{2}{c}{Val14 Set} & \multicolumn{2}{c}{TestHard Set} & \multicolumn{2}{c}{Val4k Set} & \multicolumn{2}{c}{Test4k Set} & InterPlan \\
\multicolumn{1}{l}{\bf Methods} & NR $\uparrow$ & R $\uparrow$ & NR $\uparrow$ & R $\uparrow$ & NR $\uparrow$ & R $\uparrow$  & NR $\uparrow$ & R $\uparrow$ & Score $\uparrow$ \\ 
\midrule
\textcolor{gray}{Expert (log replay)} & \textcolor{gray}{94} & \textcolor{gray}{80} & \textcolor{gray}{85.96} & \textcolor{gray}{68.80} & \textcolor{gray}{93.08} & \textcolor{gray}{76.91} & \textcolor{gray}{93.12} & \textcolor{gray}{75.01} &  \textcolor{gray}{14.76}\\
IDM \cite{treiber2000congested} & 75.59 & 77.32 & 56.16 & 62.26 & 72.52 & 76.03 & 72.45 & 74.10 & - \\
GameFormer \cite{Huang_2023_ICCV} & 82.95 & 83.88 & 66.59 & 68.83 & - & -& -& - & -\\
PDM-Hybrid \cite{dauner2023parting} & 92.84 & 92.12 & 65.07 & 75.18 & 90.31 & 90.62 & 90.65 & 90.13& 41.61 \\
PlanTF \cite{cheng2024rethinking} & 84.72  & 76.25 & 72.59 & 60.62 & 80.52 & 70.84 & 80.90 & 70.62 & 30.53 \\
DTPP \cite{huang2023dtpp} & 71.66 & 66.09 & 60.11 & 63.66 & 68.43 & 65.24 & 71.98 & 70.20 & 30.32 \\
STR-16m \cite{sun2023large} &45.06  &49.69 & 27.59&36.13 & 42.27& 45.16 & 39.66& 39.62& - \\
STR2-CPKS-100m (Ours) & 92.79 & 92.18 & 74.25 & 78.40 & 90.45 &91.14 &90.66 &90.25 & 42.13 \\
STR2-CKS-800m (Ours)  & 92.32 & 92.12 & 74.21 & 78.58 & 90.64 & 91.15& 90.35 & 90.37& 44.77 \\
STR2-CPKS-800m (Ours) & \bf 93.91 & \bf 92.51 & \bf 77.54 & \bf 82.02 & \bf 91.41& \bf 91.53& \bf 92.14& \bf 91.38& \bf 46.03 \\
\bottomrule
\end{tabular}
\end{center}
\vspace{-2mm}
\end{table*}

\begin{table*}[t]
\caption{Performance comparison on testHard Set with details of the closed-loop reactive simulations. Higher scores indicate better performance. Col. refers to No at-fault Collisions. Drivable refers to Drivable area compliance. Direction refers to Driving direction compliance. Making Prog. refers to Ego progress along the expert’s route ratio. TTC refers to Time to Collision (TTC) within bound. Speed Limit refers to Speed limit compliance. Prog. on Route refers to Ego progress along the expert’s route ratio.}
\vspace{-2mm}
\label{table:closedloop_detail}
\begin{center}
\begin{tabular}{l*{9}{c}}
\toprule
\bf Methods & Score & Col. & Drivable & Direction & Making Prog. & TTC &Speed Limit & Prog. on Route & Comfort \\
\midrule
\textcolor{gray}{Expert (log replay)} & \textcolor{gray}{68.80} & \textcolor{gray}{77.02} & \textcolor{gray}{95.96} & \textcolor{gray}{98.16} & \textcolor{gray}{100.00} & \textcolor{gray}{69.85} & \textcolor{gray}{94.12} & \textcolor{gray}{98.48} & \textcolor{gray}{99.26} \\
GameFormer\cite{Huang_2023_ICCV} & 68.83 & - & - & - & - & - & - & - & - \\
PlanTF \cite{cheng2024rethinking} & 60.62 & 90.07 & 94.85 & 97.98 & 80.51 & 85.66 & 97.97 & 65.22 & 92.28 \\
PDM-Hybrid \cite{dauner2023parting} & 75.18 & 95.22 & 95.58 & \bf 99.08 & 93.38 & 84.19 & 99.53 & 75.47 & 83.45 \\
Hoplan \cite{hu2023imitation} & 75.06 & 89.33 & 94.85 & 96.13 & \bf 97.05 & 80.51 & 95.28 & \bf 85.02 & \bf 98.52 \\
GUMP hybrid \cite{hu2024solving} & 77.77 & 94.36 & \bf98.98 & 98.95 & 94.41 & 87.46 & 97.51 & 77.08 & 79.84 \\
STR2-CPKS-100m (Ours) & 78.40 & 96.51 & 96.32 & 98.90 & 94.49 & 85.29 &  \bf 99.70 & 77.91 & 83.46\\
STR2-CKS-800m (Ours) & 78.58 & 96.32 & 96.69 & 98.90 & 94.49 & 84.56 & \bf 99.70 & 79.29 & 86.02 \\
STR2-CPKS-800m (Ours) & \bf 82.02 & \bf 97.98 & 96.69 & \bf99.08 & 94.12 & \bf87.87 & 99.27 & 78.86 & 95.59 \\
\bottomrule
\end{tabular}
\end{center}
\vspace{-6mm}
\end{table*}

Learning-based motion planners promise better generalization performance than rule-based planners with enough data. Surprisingly, previous methods filter and only train with a subset, about 1 million scenarios, of all available training samples. Following \cite{sun2023large}, we remove scenario type filters leading to over 7 million NuPlan scenarios for training. 
For the test set, we sample 100 scenarios for each type (14 different types in total provided for testing by the NuPlan devkit) leading to a total of 4790 scenarios for Val4k and 4639 scenarios for Test4k.
We also test and compare all methods on the val14 (for better comparison) and testHard (testing generality only on hard cases) subsets. For the closed-loop simulations, reactive surrounding agents challenge the generality of learning-based methods more with out-of-distribution road users' trajectories in responses. Finally, we compare different methods with a more dynamic and complicated testset, InterPlan \cite{hallgarten2024can}.

\subsection{Data Preprocess and Rasterization}
\textbf{Preprocess and Caching.} We make several optimizations to save disk usage: 1. we convert all data from float64 to float32; 2. we merge each crowd of pedestrians into one large shape; 3. we delete unused information like speed and accelerations of these road users. We sample scenarios with an interval of 1 second, producing 7 million scenarios for training. Not like \cite{dauner2023parting}, we do not make route corrections on the training dataset. We filter total static scenarios, in which the ego vehicle keeps static for the whole 10 seconds (2 seconds in the past and 8 seconds in the future).

\textbf{Rasterization.} For rasterizations, we keep each type of road shape and each type of road user in a separate channel, represented as boolean values of occupancies. The total number of channels is 34. We rasterize the map and agents into two 224x224 images. One covers a smaller range for slow but delicate movements and one covers a larger range for fast movements. The fast OpenCV library proceeds the rasterization process during training and testing without the need for caching features ahead.

\subsection{StateTransformer-2}
In this section, we explain model designs of \modelnamespace in detail. All encoded embeddings are formulated into one sequence for the decoder-only MoE backbone, shown in \ref{fig:model}. Compared to STR \cite{sun2023large}, we eliminate the diffusion-based decoder for better efficiency, scalability, and overall performance.

\textbf{ViT encoder.} We employ a decoder-only ViT image encoder for better scalability and performance, which consists of a stacked 12 layers of Transformers. Rasterized images are sliced into 16 patches. We apply no attention dropouts on the ViT encoder. We select GeLU \cite{hendrycks2016gaussian} as the activation function for the ViT encoder.

\begin{table*}[t]
\caption{Performance comparison with various state-of-the-arts with NuPlan Open-Loop Metrics.}
\vspace{-2mm}
\label{table:openloop}
\begin{center}
\begin{tabular}{lccccccccc}
\toprule
 & \multicolumn{3}{c}{Val4k Set} & \multicolumn{3}{c}{Test4k Set} & \multicolumn{3}{c}{Val14 Set} \\
\multicolumn{1}{l}{\bf Methods} & OLS $\uparrow$ & 8sADE $\downarrow$ & 8sFDE $\downarrow$ & OLS $\uparrow$ & 8sADE $\downarrow$ & 8sFDE $\downarrow$ & OLS $\uparrow$ & 8sADE $\downarrow$ & 8sFDE $\downarrow$ \\ 
\midrule
PlanCNN \cite{renz2022plant} &-  &- &- &- &- & - & 64 & 2.468 & 5.936  \\
PDM-Hybrid \cite{dauner2023parting} & 84.06 & 2.435 & 5.202 & 82.01 & 2.618 & 5.546 & 84 & 2.382 & 5.068 \\
PlanTF \cite{cheng2024rethinking} & 88.59 & 1.774 & \bf 3.892 & 87.30 & 1.855 & \bf 4.042 & 89.18 & 1.697 & \bf 3.714  \\
DTPP \cite{huang2023dtpp} & 65.15  &4.196 &9.231 &64.18 &4.117 &9.181 & 67.33 & 4.088 & 8.846 \\
STR-124m \cite{sun2023large}  & 81.88& 1.939& 4.968&82.68 &2.003 & 4.972& 88.0  & 1.777 & 4.515 \\
STR2-CKS-800m (Ours) & \bf 90.07 & \bf 1.473 &4.124 & \bf 89.12 & \bf 1.537 & 4.269& \bf 89.2 & \bf 1.496 & 4.210 \\
\bottomrule
\end{tabular}
\end{center}
\vspace{-4mm}
\end{table*}

\textbf{Mixture-of-Expert.} 
Language modeling tasks also require models to learn and balance through complicated and often stochastically controversial rewards from expert data. 
Inspired by the generalization results of the MoE models on the language modeling tasks, we replace the GPT-2 backbone \cite{radford2019language} with an MoE backbone for sequence modeling. The MoE backbone, inspired by \cite{fedus2022review, fedus2022switch}, is based on the Transformer architecture \cite{vaswani2017attention} with modification of changing the feed-forward blocks to Mixture-of-Expert layers. The MoE layers provide much better memory efficiency through specialized kernels and Expert Parallelism (EP). We also utilize the Flash Attention 2 \cite{dao2023flashattention} and the Data Parallelism (DP) for better training efficiency. The backbone architecture parameters for different sizes are summarized in Table \ref{MoE_Backbone}.

\textbf{Autoregression and Proposal.} We add a proposal embedding feature in the generation sequence for modality classifications with a Cross-Entropy loss, similar to the implementations in \cite{sun2023large, shi2022motion} on the motion prediction task with the Waymo Open Motion Dataset (WOMD) \cite{ettinger2021large}. 
Following \cite{li2024hydra}, we extract 512 proposals with K-Means clustering with a mini-batch of 1000 from 0.7 million randomly selected dynamically feasible trajectories by their spatial-temporal distances. Each normalized trajectory includes 80 points (x, y, and yaw) for the future 8 seconds.

\textbf{Sampling and Scoring}: 
In NuPlan closed-loop simulations, the ego vehicle is controlled by a Linear–quadratic regulator (LQR) controller producing tracking errors to bring a systematic bias from the model's output and the final trajectory to compute the metrics. Inspired by \cite{dauner2023parting}, we apply similar sampling and scoring methods to enhance the model's output.
Specifically, we apply lateral offsets of [-1 m, 0 m, 1 m] to the model's output trajectory with the route centerline, resulting in four lateral candidate paths. 
We set 5 different target speeds for diverse lateral sampling of [0.2, 0.4, 0.6, 0.8, 1.0] of the maximal speed limit.
With the original model's output, 21 trajectory candidates are sent to an LQR controller. The 21 tracking results are then scored as in \cite{dauner2023parting}.

\subsection{Training and Testing Settings}

We train both the ViT encoder and the MoE backbone from scratch. We train \modelnamespace with a batch size of 16 on 8 H20 (96GB) Nvidia GPUs for 20 epochs for the STR2-CPKS-800m model. We train \modelnamespace with a batch size of 64 on 8 3090 Nvidia GPUs for 20 epochs for the STR2-CPKS-100m model. We run Open-Loop and Closed-Loop simulations with a batch size of 50. The Val4k and Test4k simulations take about 5 hours to run closed-loop simulations on each. We use a Cosine-Restart learning rate scheduler, restarting a cosine LR schedule after each epoch, to avoid overfitting at a local minimum. All \modelnamespace are trained with bfloat16 for training efficiency. 

\subsection{Scaling on the LiAuto Dataset}
The LiAuto dataset is an industrial-level extra-large real-world driving dataset. 
This dataset includes a lane-level navigation map and tracking results from a sensor setting of 7 RGB cameras, 1 LiDAR, and 1 millimeter-wave radars (MMWR).
The same dataset is used for training models which are later deployed on a fleet with a size of over 900,000 different models of vehicles.
We select urban driving scenarios collected from the last 6 months without any human labeling.
We filter the scenarios with the wrong navigation routes, not matched with actual future driving trajectories. 
Finally, we reformulate all driving logs into training and testing samples with a length of 10 seconds, 2 seconds in the past, and 8 seconds in the future.
The final training dataset has over 1 billion training samples.

\begin{table}[t]
\caption{Scaling study on the impact of model parameters with \modelname-CKS evaluated on NuPlan Open-Loop Metrics.}
\label{table:open_loop_ablation}
\begin{center}
\vspace{-2mm}
    \begin{tabular}{lcccc}
    \toprule
    \textbf{STR} & Val14-OLS & Val14-8sADE & Val14-8sFDE & MR \\
    \midrule
    STR2-100m & 88.56 & 1.55 & 4.47 & 3.05\% \\
    STR2-800m & 89.02 & 1.48 & 4.20 & 2.87\% \\
    STR2-1b   & 89.74 & 1.46 & 4.13 & 2.15\% \\
    \bottomrule
    \end{tabular}
\end{center}
\vspace{-8mm}
\end{table}

\begin{figure*}[t]
    \centering
    \includegraphics[trim=0cm 5.2cm 0cm 5.2cm, width=0.99\textwidth]{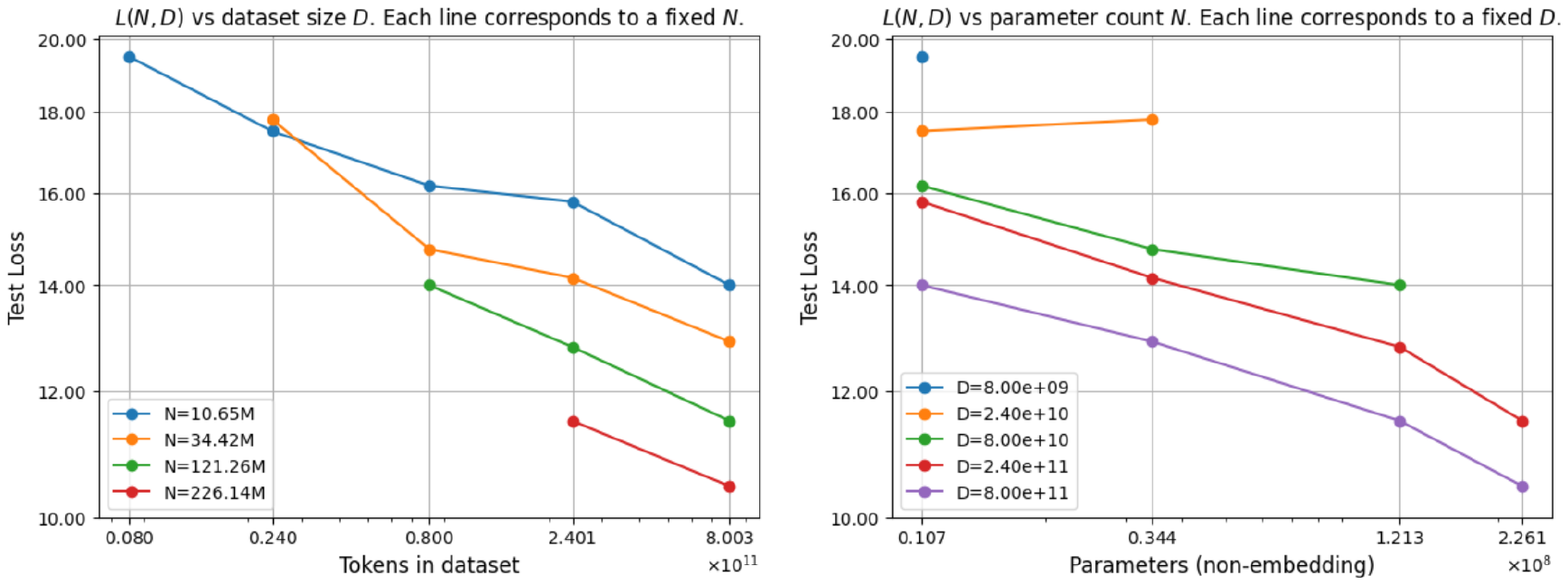}
    \caption{Scaling results with the size of the training dataset, counted as the number of tokens $D$ in the left and scaling results with model parameters $N$ in the right. All axes are logarithmically scaled. }
    \label{fig:scalinglaw}
\vspace{-4mm}
\end{figure*} 


\section{Results}

In this section, we present, discuss, and analyze the experiment results of \modelnamespace on the open NuPlan dataset and the large LiAuto dataset. Despite learning from more challenging raster representations, \modelnamespace achieves comparable open-loop performance and beats other SOTA methods on the Closed-Loop metrics with better overall scores, suggesting better generalization in learning and balancing explicit driving rewards. Additionally, we present a comprehensive scaling analysis of training on the LiAuto dataset, implying strong scalability across extra-large datasets of up to 1 billion training samples.

\subsection{Open-Loop Performance}

The open-loop evaluation loss, or 8sADE can be considered a direct indicator of the learning abilities of different models.
As shown in Table \ref{table:openloop}, \modelnamespace outperforms the other methods on 8sADE accuracy and is comparable on other Open-loop metrics, suggesting strong fitting ability. 
Additionally, we also evaluate \modelnamespace by training with different sizes, in response to the trainable model parameters. As shown in \ref{table:open_loop_ablation}, larger models tend to fit the dataset distribution better than smaller ones, indicating great scalability along model sizes.


\subsection{Closed-Loop Performance}
\label{section:closedloop}
Closed-loop performance is the spotlight metric for benchmarking different motion planners. We test a wide spectrum of planners \footnote{No GameFormer \cite{huang2023gameformer} checkpoint available for re-evaluations.} with various test sets. To further challenge the generalization ability, we test these methods against the novel closed-loop benchmark, InterPlan \cite{hallgarten2024can}. InterPlan constructs out-of-distribution testing scenarios like dealing with a crush site in the middle of the road. As shown in Table \ref{table:closedloop}, \modelnamespace outperforms the other methods on all 4 test sets, including the most popular Val14 set \cite{dauner2023parting}, the hard-cases TestHard set \cite{cheng2024rethinking}, and two larger scale Val4k and Test4k sets. Details of each metric are shown in Table \ref{table:closedloop_detail}

Beyond numerical results, we uncover several insights through comparison. Firstly, in the reactive closed-loop simulations, other road users will always yield to the ego vehicle with each conflict. This means the overall score should always be higher on reactive simulations than on non-reactive simulations with the same test set due to fewer collisions. Surprisingly, we discover a performance drop of the PlanTF \cite{cheng2024rethinking} and the DTPP \cite{huang2023dtpp} when tested with reactive surrounding agents. These drops indicate inferior generalizations of these planners than the others because they are more likely to be distracted by other road users' out-of-distribution reactions due to a severe overfit to the original expert demonstrations, probably a result of causality confusion during training. 
Although they spot the problem of causality confusion on the ego past states and fix it by state dropouts, the same problem remains on the past trajectory of the other agents. 
This conclusion is double-confirmed in the InterPlan testings with similar performance drops. On the other hand, \modelnamespace generalizes better than previous methods on reactive simulations. Additionally, we discover a significantly better generalization performance than other methods on the testHard and InterPlan test sets. The testHard set contains filtered in-distribution few-shot cases, like negotiating with other road users, and complicated cases, like balancing multiple conflicts. The InterPlan contains zero-shot cases, like driving through construction areas. \modelnamespace suffers the least performance deterioration indicating the best generalization abilities by scaling the MoE backbone. Finally, \modelnamespace significantly outperforms STR, the same model structure with a GPT2 backbone, with the closed-loop simulations.

\subsection{Scaling to Billions on LiAuto Dataset}

We study the scalability of our method by evaluating performance over the size of the dataset and model parameters. Results show that they follow a log-log relationship \cite{kaplan2020scaling} with the evaluation/test loss, producing outstanding generalization performance up to billions of training samples. Following previous methods to test the scalability, we examine the converged test loss, which is the L2 loss for motion planning with \modelname-CKS models. We test \modelnamespace across 3 magnitudes on the size of the dataset (up to 1 billion) and over 2 magnitudes on the size of the models (up to 300M). As shown in Fig. \ref{fig:scalinglaw}, the left image illustrates the relationship between the test loss L and the dataset size D, denoted as L(D), with varying model sizes N. The right image shows the relationship between the test loss L and the number of parameters N, denoted as L(N), with varying dataset sizes D. Specifically, we observe that $L(N)$ can be fitted with $L(N)=(N_c/N)^{\alpha_N}$ where $N$ represents the number of trainable model parameters excluding encoding and positional embeddings and $L$ denotes the L2 loss, and $\alpha_N$ is the power-law exponent characterizing the scaling behavior of the loss.

\section{CONCLUSIONS}

In this paper, we propose a scalable and strong model, \modelname, for generalization on motion planning for autonomous driving. We demonstrated that previous methods often rely on overly complex designs that may hinder scalability when applied to large-scale datasets. Through extensive benchmarking on the NuPlan dataset and industrial-scale datasets, we showed that our decoder-only MoE architecture significantly improves generalization across diverse driving scenarios, including challenging out-of-distribution zero-shot cases.
Our experiments confirm that MoE models when trained at scale, can achieve superior planning performance over more complex architectures and training paradigms by balancing multiple explicit rewards. 


\textbf{Limitation and future work.} 
We leave comprehensive scaling analysis on the LiAuto dataset with larger models as future works due to large computation requirements. Testing the interaction-intensive scenarios with the NuPlan reactive simulations suffers from limited environment agents' (vehicles only) behavior simulations.  We leave more tests with more advanced simulators as future works.
Another limitation of \modelnamespace is the inference time. Larger models might generate slower. We leave speeding up the inference time on edge computing resources as future works.

\addtolength{\textheight}{-1cm}   





\section*{ACKNOWLEDGMENT}

We would like to thank Yujian Li for speeding up the NuPlan simulator, Qinghui Zhao for retesting previous methods, Tianyu Zhang for CPKS model training, Yue Wang for trajectory sampling, Weixing Cai, Yangang Zou, and Wentao Li for simulation visualization and analysis.


\bibliographystyle{IEEEtran}
\bibliography{references}

\end{document}